# Deep Smoke Segmentation

Feiniu Yuan*, *Senior Member, IEEE*, Lin Zhang, Xue Xia, Boyang Wan, Qinghua Huang and Xuelong Li, *IEEE Fellow*

*Abstract*—Inspired by the recent success of fully convolutional networks (FCN) in semantic segmentation, we propose a deep smoke segmentation network to infer high quality segmentation masks from blurry smoke images. To overcome large variations in texture, color and shape of smoke appearance, we divide the proposed network into a coarse path and a fine path. The first path is an encoder-decoder FCN with skip structures, which extracts global context information of smoke and accordingly generates a coarse segmentation mask. To retain fine spatial details of smoke, the second path is also designed as an encoder-decoder FCN with skip structures, but it is shallower than the first path network. Finally, we propose a very small network containing only add, convolution and activation layers to fuse the results of the two paths. Thus, we can easily train the proposed network end to end for simultaneous optimization of network parameters. To avoid the difficulty in manually labelling fuzzy smoke objects, we propose a method to generate synthetic smoke images. According to results of our deep segmentation method, we can easily and accurately perform smoke detection from videos. Experiments on three synthetic smoke datasets and a realistic smoke dataset show that our method achieves much better performance than state-of-the-art segmentation algorithms based on FCNs. Test results of our method on videos are also appealing.

## I. INTRODUCTION

As early as 1979, smoke detection had been proposed for solving the fire safety problem [1]. In the past decades, many researchers have been committed to smoke detection for reducing damages caused by fire, since smoke can provide earlier clues for fire alarms than flame. Strictly, smoke detection can be divided into two categories. The first category is to merely judge whether there is smoke in an image or not. It is also known as whole image smoke recognition. The other one is not only to identify whether there is smoke, but also to indicate the accurate location of smoke. This category may be smoke detection or smoke segmentation.

Smoke segmentation requires accurate seperation of smoke from background at pixel levels, while smoke detection [2][3][4][5][6] only needs to give a rough position of smoke, which is usually marked with a bounding box. Smoke segmentation often needs to generate a mask with detailed edges, which involves object classification, localization and boundary delineation. Therefore, smoke segmentation is a far more difficult task than smoke recognition. Traditional smoke segmentation methods mainly use various features of smoke, such as color [7][8][9][10], shape [11], motion [12]. However, traditional features are very difficult to design, define or choose due to large variations of smoke appearance, resulting in very poor segmentation. In addition, most of smoke segmentation methods are based on videos [7][9][11][12]. But in [12], Tian [13] pointed out that video-based methods are very unstable in the cases of bad weather and limited infrastructure resources. Therefore, smoke segmentation from static images play a very important role in visual smoke detection.

Fully Convolutional Networks (FCN) were first used in semantic segmentation in [14], which implemented an end-to-end pixel-wise prediction. Semantic segmentation based on FCNs can accept an input image of arbitrary size, and utilize a set of convolutional layers, non-linear activation functions, pooling and upsampling layers to output a predicted map for the input image. The major operation used in FCNs is convolution, which has the advantage of computational simplicity, effectiveness and local weight sharings. FCNs can perform an end-to-end semantic segmentation quickly and accurately [15]. In recent years, FCNs have achieved a lot of significant results in object segmentation tasks [15][16][17][18][19][20]. In addition, the performance of FCNs on whole image classification [21][22][23][24] is also very excellent.

Inspired by the recent success of FCNs, we present a two-path encoder-decoder network based on a FCN architecture to acquire semantic segmentation of smoke images. Each pixel is classified as smoke or non-smoke labels. Compared to segmentation of general objects, it is a more challenging task to accurately segment fuzzy and translucent smoke objects from a single image. To overcome the fuzzy and translucent issue, we propose to divide smoke segmentation into two paths: coarse

Manuscript received Aug 28, 2018. This work was partially supported by the National Natural Science Foundation of China (61862029), and the Science Technology Application Projects of Jiangxi Province (KJLD12066, GJJ170317).

Feiniu Yuan is with College of Information, Mechanical and Electrical Engineering, Shanghai Normal University, Shanghai 201418, China. (e-mail: yfn@shnu.edu.cn)

Lin Zhang, Xue Xia and Boyang Wan are with the School of Information Technology, Jiangxi University of Finance and Economics, Nanchang, Jiangxi, 330032, China. Lin Zhang is also with School of Mathematics and Computer Science, Jiangxi Science and Technology Normal University, Nanchang 330038, China.

Qinghua Huang is with School of Mechanical Engineering, and Center for OPTical IMagery Analysis and Learning (OPTIMAL), Northwestern Polytechnical University, Xi'an 710072, China. (e-mail: qhhuang@nwpu.edu.cn)

Xuelong Li is with Xi'an Institute of Optics and Precision Mechanics, Chinese Academy of Sciences, Xi'an 710119, China. (e-mail: xuelong_li@ieee.org)

*Corresponding authors.

and fine segmentations. In addition, due to the lens of cameras, smoke has large scale variations, so it is also critical to solve the multi-scale problem. There are two main ways to deal with this issue [25]. The first way is to feed images with different sizes to the network, but it is time consuming. The second one is to fuse feature maps from different layers using skip operations, hence it is effective and efficient. In our method, we follow the second way to address the scale issue.

To the best of our knowledge, this is the first time to design an end-to-end two-path encoder-decoder FCN for visual smoke segmentation. To speed up training, we use the last three blocks of the first five blocks in the VGG16 network [26] for the first path network. Although deep image matting [27] uses a two-stage network that are a little similar to our two-path network, our method is obviously different. The first difference is that the refinement network of our method is also an encoder-decoder with skip structures, which is much effective and powerful than that of [27]. The second one is that our method is trained end-to-end instead of three-stage training techniques used in [27]. Thus, we can simultaneously optimize our proposed network. The main contributions of this paper are summarized as follows.

- First, we propose an end-to-end two-path FCN by fusing coarse and fine segmentation of smoke images. The first path network is a deep encoder-decoder FCN to acquire global information while the second one is just a shallow encoder-decoder for local fine information. Deeper layers of the first network are used to capture global context information for coarse location of smoke, while shallower layers of the second network can obtain local fine information to enrich the details of smoke. Hence, combination of the two networks finally produces more accurate segmentation results.

- Second, we propose a very small network containing only add, convolution and activation layers to fuse the results of the two paths. Thus, we can train the proposed network end to end for simultaneous optimization of network parameters. Feature maps from different layers with skip structures in both deep and shallow networks carry multi-scale, coarse and fine information, so the fusion network can efficiently improve accuracy.

- Third, we propose a method to virtually generate novel datasets for smoke binary segmentation. The motivation of this work is that currently there is lack of unified datasets in the field of smoke segmentation for training and fair comparisons. It is very difficult and time-consuming to manually create accurate ground truths for realistic smoke images. Our synthesizing method can avoid the difficulty.

The remainder of the paper is organized as follows. Related work on FCNs for object segmentation is given in Section II-A, and smoke segmentation in Section II-B. The generation method of smoke segmentation datasets is introduced in Section III. A detailed description of the proposed network is presented in Section IV. In Section V, a large number of experimental results and analysis are shown to evaluate the performance of the proposed method. Conclusions are drawn in the last section.

## II. RELATED WORK

### A. Fully convolutional network for segmentation

Long *et al*. [14] first used FCNs to truly implement an end-to-end semantic segmentation, which has a pioneering significance for object segmentation. The authors achieved FCNs by converting the last fully connected layers into 1*1 convolutions. They also added skip structures in the network to fuse outputs of different layers and obtained high quality segmentation results. The reason why the authors use skip layers is that skip structures provide finer spatial information since posterior layers of an encoding network lose a lot of spatial details due to multiple pooling operations. Upsampling the outputs of these posterior layers produces very coarse and blocky segmentation. In addition, in the training process, the authors directly used the parameters of VGG16 pre-trained on the ImageNet [28] to initialize the network, and then adopted other datasets to fine-tune the network, which can speed up the convergence of networks.

Recently, a deep convolutional encoder-decoder network termed SegNet [29] was presented. The number of trainable parameters of FCNs in [14] is too large, leading to the difficulty in training end-to-end for a relevant task. Therefore, a stage-wise training technique is utilized in SegNet, which differs from FCNs in [14] in several aspects. First, SegNet adds batch normalization [30] after each convolutional layer and it is a smaller network and easier to train by removing the fully connected layers of VGG16. Second, the decoding stage of SegNet contains a series of convolution and upsampling operations, and there is a consistent one-to-one match between encoding and decoding stages, while FCNs in [14] have only upsampling operations in the decoding stage. Finally, instead of using skip structures to obtain detailed information, SegNet uses unpooling operations to capture and store boundary information of the encoder feature maps by saving the max-pooling indices.

It is worth mentioning that U-Net [17] is more like a combination of the above two methods. The U-Net includes both skip structures and a decoding stage that is symmetrical to its encoding network. The major difference between U-Net and the above two methods is that U-Net transfers the feature maps of an encoder to its corresponding decoder and then concatenates them by copying and cropping operations.

Some methods use skip structures that are similar to FCNs in [14]. In [31], the authors proposed a Text-Block FCN by uniformly upsampling the output of each convolutional stage to the same size and then adding all the resized outputs together to obtain the final feature map. Caelles *et al*. [18] proposed a two-stream FCN architecture, which adopted an exactly similar structure including skip connections, but with different loss, to obtain the foreground and contour of objects. Luo *et al*. [32] proposed a 4×5 grid CNN network containing a large number of skip operations to combine local and global information. Hou *et al*. [19] introduced short connections on the basis of skip structures that provide rich multi-scale information.

Most of the above methods are all based on VGG16, and other methods [33][34][35] use ResNet [36]. Jain *et al*. [33]

designed a two-stream FCN by combining ResNet-101 with dilated convolutions for fusion of motion and appearance. Lin *et al*. [34] and Peng *et al*. [35] used the residual structures of ResNet to refine object boundaries.

Another way to improve performance is to combine FCNs with graphical models. The most representative method is a series of networks of Deeplab [37][38]. Deeplab v1 [37] introduces dilated convolutions to expend receptive fields, and connects a fully connected conditional random field (CRF) to the end of FCNs for further refinement of segmentation results. Deeplab v2 [38] adds the atrous spatial pyramid pooling (ASPP) on the basis of [37] and replaces VGG16 by ResNet101 [36]. However, the main difference between deeplab v3 [39] and the other two deeplab methods is that the fully connected CRF is abandoned. Furthermore, deeplab v3 has improved ASPP and introduces multiple grids.

Different with previous work, we propose a two-path encoder-decoder FCN by combining skip and encoder-decoder structures for smoke segmentation. As far as we know, this is the first paper to adopt skip layers and two paths in a single end-to-end network for smoke segmentation.

### *B. Smoke segmentation*

There are few literatures for semantic segmentation of smoke. Most of smoke segmentation methods are mainly based on the color characteristics of smoke [7][8][11]. In addition, rough sets and color features [9][10] are combined for smoke segmentation. For example, Gaussian mixture models (GMM) are introduced [12][40] for smoke segmentation. In [9], the authors utilized Kalman filters to update background to exclude objects with color similar to smoke, and then separated smoke by the roughness distribution of smoke colors in RGB space. Zhang *et al*. [10] used the red channel of RGB colors to construct the roughness histogram for smoke segmentation. Hu *et al*. [40] removed the long-term unmatched expired Gaussian components to reduce background modeling time. Jia *et al*. [12] used saliency maps for smoke segmentation, and at the same time adopted a motion estimation function calculated with GMMs to estimate salient regions to obtain suspected smoke regions.

Another representative method [12] is obviously different from the aforementioned methods. The method is based on the atmospheric scattering model. Two dictionaries are first trained for smoke and background, respectively. By solving the sparse representation problem, an input image is densely classified into quasi-smoke and quasi-background components. Then, image matting techniques and GMMs are introduced for separating smoke from background according to initial detection results.

To the best of our knowledge, there is currently no literature that uses deep learning for smoke segmentation. Inspired by the achievement of FCNs in traditional object segmentation, we introduce FCNs into segmentation of fuzzy smoke. According to the description in previous sections, we find that most algorithms widely use pre-trained parameters of VGG16 and skip structures for performance improvement. Moreover, in addition to Text-Block FCN, and deeplab v1 and v2, other methods generally accomplish object segmentation in a single stage. However, we also find that proper fine-tuning helps with performance improvement. To merge the advantages of the above methods, in this paper we simultaneously combine two-path encoder-decoder FCNs, skip structures and decoding structures to propose a novel FCN for smoke segmentation.

### III. DATASET GENERATION FOR SMOKE BINARY SEGMENTATION

One of the great obstacles to smoke segmentation based on deep learning is the lack of adequate annotated training data. Although we can easily acquire a huge number of smoke images, it is extremely time-consuming, boring and difficult to manually segment smoke objects from an image since smoke has fuzzy edges and translucent property.

As far as we know, there is no image dataset for smoke segmentation. It is of great significance to create image datasets for smoke segmentation in both research and industry communities of visual smoke detection. We first used computer graphics to virtually generate 8162 pure smoke images, denoted as a PureSmoke dataset. Then, we adopted a linear color composition method to synthesize a random background image and a pure smoke image to construct a training dataset and three test datasets. Each test dataset has 1000 synthesized smoke images for comparisons.

Some pure smoke samples are shown in Fig. 1. Each image is an RGBA image containing 4 channels, i.e., three RGB color channels (**S**) and one alpha channel ($\alpha$). We use the linear color composition of a pure smoke image (**S** and $\alpha$) and a background RGB image **B** to generate a new smoke RGB image **I**, which is expressed as follows:

$$\mathbf{I} = (1-\alpha)\mathbf{B} + \alpha\mathbf{S} \qquad (1)$$

where the alpha $\alpha$ is actually a blending parameter in the range [0, 1] representing the transparency of smoke.

To make the training samples diverse, we use the Places365-Standard dataset [41] as background images. In the synthesis process, we randomly select a background image from the Places365-Standard dataset and a pure smoke image from the PureSmoke dataset, then combine the two images to generate a synthesized smoke image. To further perform data augment of smoke images, we change the alpha values of pure smoke images to control the concentrations of smoke images, which is expressed as:

$$I_R = (1-\alpha\beta)B_R + \alpha\beta S_R \quad (2)$$

$$I_G = (1-\alpha\beta)B_G + \alpha\beta S_G \quad (3)$$

$$I_B = (1-\alpha\beta)B_B + \alpha\beta S_B \quad (4)$$

where $\beta$ is a linear coefficient with range (0,1) used to control the concentration of smoke, $I_R$, $I_G$ and $I_B$ are the red, green and blue channels of a pixel in an observed smoke RGB image **I**, respectively, $S_R$, $S_G$ and $S_B$ are respectively the red, green and blue channels of the pixel in a pure smoke RGBA image (**S** and $\alpha$), and $B_R$, $B_G$ and $B_B$ are respectively the red, green and blue channels of the same pixel in a background RGB image **B**.

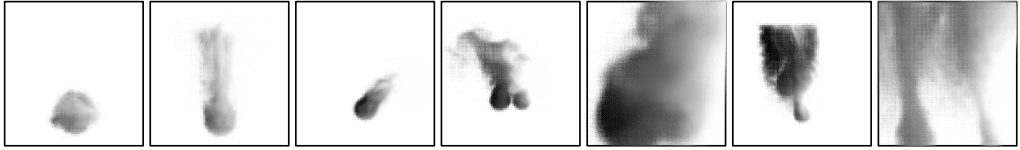

Fig. 1.    Pure smoke images generated by computer graphics

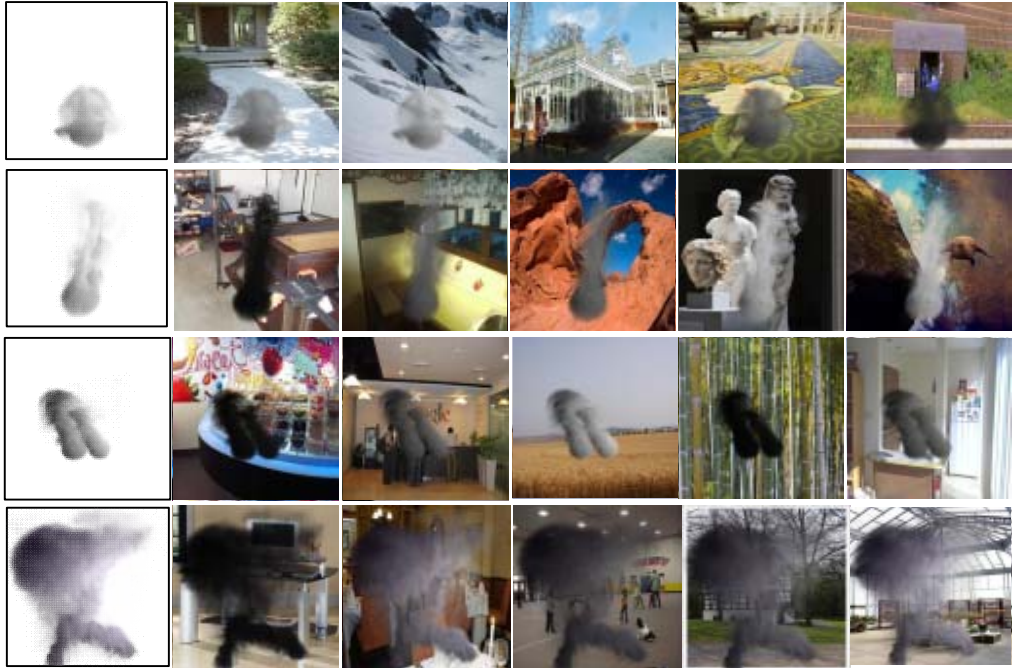

Fig. 2.    Pure smoke images and synthesized images

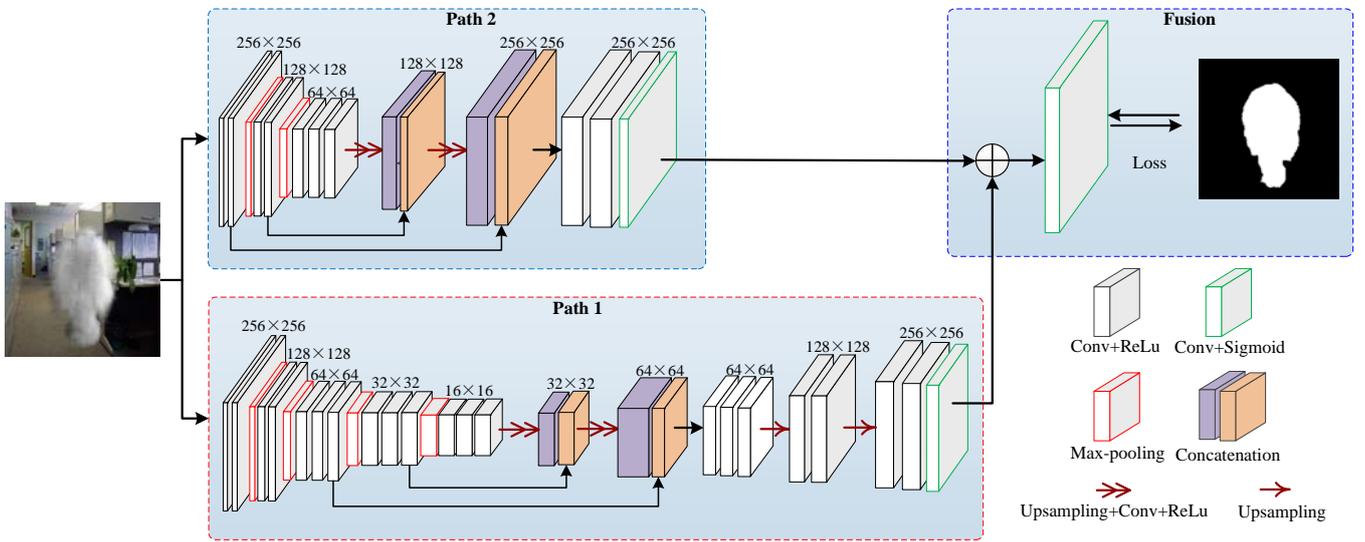

Fig. 3.    Our network structure with two paths

In this way, we can easily generate a large number of training samples. Fig. 2 shows some training images for smoke segmentation. Our synthesizing method avoids the difficulty in manually annotating ground truths of smoke images. We threshold the alpha channel of a pure smoke image to generate a binary mask for smoke regions, and then the binary mask is regarded as the ground truth of a synthesized image blended by the pure smoke image and any other background image.

## IV. TWO-PATH FULLY CONVOLUTIONAL NETWORKS

Ideally, we require our deep smoke segmentation network to have the following advantages: accurate segmentation, small network size, and a fast test speed. The overall structure of the proposed method is shown in Fig. 3. This network is called a two-path FCN since it consists of two paths. The first path is a deep fully convolutional network with asymmetrical structures, which is used for global segmentation prediction. The second one is a refinement path that produces a more detailed prediction on the basis of global prediction. Then we will describe the two paths in more detail in following sections.

### A. A deep network for global context information

The first path aims at gaining global context information for generation of a coarse smoke segmentation map. The network of this path takes a single RGB image as input and produces a prediction map with the same size of the input. The detailed structure of the network is shown in the bottom row of Fig. 3. Apparently, the network is a typical encoder-decoder FCN. To speed up training, there are many segmentation algorithms [14][31][37][38][39] that adopt the convolutional blocks of the VGG16 network [26] to achieve remarkable results. We also utilize the first five blocks of the VGG 16 network as the basis of the encoding phase of path 1, which contains 13 convolutional layers and 4 max-pooling layers, and we remove the fully connected layers to reduce the number of trainable parameters.

Table 1. The hyper-parameters of path 1

| Encoding phase | | |
|---|---|---|
| Block1 | Conv+ReLu | 3*3*64 |
| | Conv+ReLu | 3*3*64 |
| | Max-pooling | 2*2 |
| Block2 | Conv+ReLu | 3*3*128 |
| | Conv+ReLu | 3*3*128 |
| | Max-pooling | 2*2 |
| Block3 | Conv+ReLu | 3*3*256 |
| | Conv+ReLu | 3*3*256 |
| | Conv+ReLu | 3*3*256 |
| | Max-pooling | 2*2 |
| Block4 | Conv+ReLu | 3*3*512 |
| | Conv+ReLu | 3*3*512 |
| | Conv+ReLu | 3*3*512 |
| | Max-pooling | 2*2 |
| Block5 | Conv+ReLu | 3*3*512 |
| | Conv+ReLu | 3*3*512 |
| | Conv+ReLu | 3*3*512 |
| Decoding phase | | |
| Block6 | Upsampling | 2*2 |
| | Conv+ReLu | 3*3*512 |
| | Concatenation | / |
| Block7 | Upsampling | 2*2 |
| | Conv+ReLu | 3*3*512 |
| | Concatenation | / |
| Block8 | Conv+ReLu | 3*3*256 |
| | Conv+ReLu | 3*3*256 |
| | Conv+ReLu | 3*3*256 |
| Block9 | Upsampling | 2*2 |
| | Conv+ReLu | 3*3*128 |
| | Conv+ReLu | 3*3*128 |
| Block10 | Upsampling | 2*2 |
| | Conv+ReLu | 3*3*64 |
| | Conv+ReLu | 3*3*64 |
| Prediction phase | | |
| | Conv+Sigmoid | 1*1*1 |

In order to obtain multi-scale features and retain detailed spatial information, we increase the network depth and add skip structures between encoding and decoding phases of the network. Zhang et al. [31] and Hou et al. [19] verified that deeper layers can capture more global information. Learned from the above idea, we propose to incorporate the last three blocks of the encoding phase into the decoding phase to increase the network depth for capturing more global information. The output feature maps of convolutional layers in the encoding phase are extracted, and then these feature maps are upsampled to the size of the feature maps of corresponding decoder layers to perform feature concatenation operations.

To further reduce the number of network parameters and improve the training speed, we use an asymmetric structure in the decoding phase. Besides two concatenation layers, the decoder network only includes 9 convolutional layers and 4 upsampling layers. Finally, a convolutional layer with a 1×1 kernel and a sigmoid activation function, which is called prediction phase, is added at the end of the decoder network to predict a coarse segmentation map for input. The detailed hyper-parameters of the network are shown in Table 1.

The first path network leverages a binary cross-entropy loss with weight decaying regularization as a loss function, which can be expressed as:

$$L(\mathbf{P}, \mathbf{G}) = -\sum_i \left( g_i \log p_i + (1 - g_i) \log(1 - p_i) \right) + \lambda \|\mathbf{W}\|_2^2 \quad (5)$$

where $p_i$ is the probability of a pixel $i$ classified as smoke in the predicted map $\mathbf{P}$, $p_i \in [0, 1]$, $g_i$ is the probability of the pixel $i$ in the corresponding ground truth map $\mathbf{G}$, and $g_i$ is equal to 1 for a smoke pixel and 0 for non-smoke.

### B. A shallow network for local fine information

As mentioned earlier, smoke is hard to segment because it has very blurry edges and translucent property. In the first path of our network, we obtain global information to generate a coarse segmentation of smoke, but we lost detailed spatial information for smoke localization. Therefore, the goal of the second path is to capture details of smoke.

Table 2. The hyper-parameters of path 2

| Encoding phase | | |
|---|---|---|
| Block1 | Conv+ReLu | 3*3*64 |
| | Conv+ReLu | 3*3*64 |
| | Max-pooling | 2*2 |
| Block2 | Conv+ReLu | 3*3*128 |
| | Conv+ReLu | 3*3*128 |
| | Max-pooling | 2*2 |
| Block3 | Conv+ReLu | 3*3*256 |
| | Conv+ReLu | 3*3*256 |
| | Conv+ReLu | 3*3*256 |
| Decoding phase | | |
| Block4 | Upsampling | 2*2 |
| | Conv+ReLu | 3*3*256 |
| | Concatenation | / |
| Block5 | Upsampling | 2*2 |
| | Conv+ReLu | 3*3*256 |
| | Concatenation | / |
| Block6 | Conv+ReLu | 3*3*64 |
| | Conv+ReLu | 3*3*64 |
| Prediction phase | | |
| | Conv+Sigmoid | 1*1*1 |

In addition to proving that deeper layers can capture more global information, Zhang *et al*. [31] and Hou *et al*. [19] also verified that shallower layers can capture rich local information and object details. According to this idea, we propose to use a shallow encoder-decoder network using the first three blocks of VGG16 with two max-pooling layers to retain more details of the input. The detailed structure of the shallow network is shown in Fig. 3. The network is just an encoder-decoder structure that includes 7 convolutional layers and 2 max-pooling layers in the encoding phase, and 4 convolutional layers and 2 upsampling layers in the decoding phase. The hyper-parameters of the second path network are given in Table 2.

Similar to the network of the first path, the second network also adopts skip structures for capturing scale information. We extract the outputs of two convolutional layers in the encoding phase, and then upsample the extracted feature maps to the size of the feature maps in corresponding decoder layers to complete concatenation operations. Finally, we obtain the detailed segmentation map of smoke through two ReLu convolutional layers and one sigmoid convolutional layer.

### C. A fusion network of two paths

The overall goal of this work is to gain an accurate smoke segmentation map. Therefore, we merge the coarse prediction result at path 1 and the detailed spatial information at path 2 to generate the final accurate result. We first add up the network outputs of the two paths, and then feed the summed segmentation map into a convolutional layer with a 1*1 kernel and a sigmoid activation function to produce the final prediction map, as shown in Table 3.

Although we have borrowed some inspirations from the method proposed by zhang *et al*. [31], our method is obviously different from the method. First, we only use the feature information of the last three blocks in VGG16, and zhang *et al*. [30] used all the first five blocks of VGG16. Second, we use upsampling and concatenation operations to replace deconvolution and summation layers [30]. Thirdly, our method includes decoding phases with multiple upsampling and convolutional layers. Fourth, we propose a shallow encoder-decoder network with skip layers to produce finer segmentation results.

Table 3. The hyper-parameters for fusion

| Final prediction phase | |
|---|---|
| Add | / |
| Conv+Sigmoid | 1*1*1 |

### V. EXPERIMENTAL RESULTS

In this section, we will describe our experimental datasets (Sec. A) and implementation details (Sec. B), evaluate our model on three synthetic test smoke datasets and one real smoke dataset in quantitative and qualitative manners (Sec. C), explain the importance of each part of the model by ablation experiments (Sec. D), and finally show the limitations of the proposed model (Sec. E).

### A. Image datasets

We created three synthetic smoke image datasets and one real smoke image dataset for comparisons. These datasets are very challenging due to large variations in texture, shape, color and scales. All test images were not used in the training process, and even the smoke backgrounds are totally different from the training samples.

Each synthetic dataset contains 1000 composited smoke images of size 256*256, which were generated in the same way as the training data described in Section III. But the background images for generation of the three synthetic test datasets were selected from CBCL StreetScenes [43], Pascal Visual Object Classes [44] and Baidu people segmentation dataset [45], respectively. For the sake of clearness, the three synthetic test datasets with different background images are named DS01, DS02 and DS03, respectively. Fig. 4 shows some synthetic smoke images and corresponding ground truths of alpha channels from the three synthetic data sets. To create the real smoke dataset, we manually collected several smoke images from the internet. Due to lack of ground truths for real images, the real image dataset was used only for qualitative analysis.

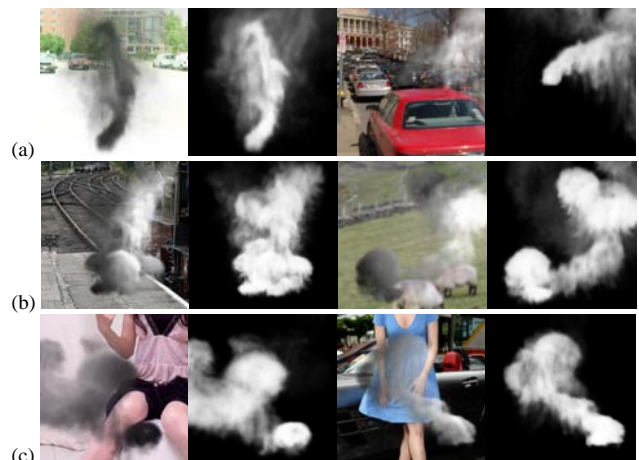

Fig. 4. Synthetic smoke images and corresponding ground truths of alpha channels from (a) DS01, (b) DS02 and (c) DS03.

### B. Implementation details

We implemented our proposed network using Tensorflow and Keras. The proposed model was trained end-to-end using a single NVIDIA GeForce GTX1080Ti with 11 GB RAM. A well-known problem with network training is the initialization of network parameters. If the initialization parameters of network are not suitable, it is easy to get the network into local optimal solutions. Related work has shown that this problem can be overcome by using pre-trained parameters of some famous networks to initialize the network. Therefore, for the encoder, 13 convolutional layers were initialized with the weights of the VGG16 network pre-trained on ImageNet. The parameters of the decoder were randomly initialized by truncated normal distributions. Since the network does not use smoke images for further training, it is unable to perform smoke segmentation. Therefore, we trained our network on the smoke training datasets with stochastic gradient descent (SGD) [42] with a fixed learning rate of 0.001, a momentum of 0.9 and a weight decay of 1e-5.

Our method can be efficiently trained end-to-end because we use the fusion network and skip structures. However, the method in [27] adopts a complicated updating strategy for training. It first updates the parameters of the first path network by removing the second path network. Once the first path is converged, it fixes the first path parameters and then updates the second path network without the first network. After the second path is converged, it fine-tunes the whole network again using the same training method. Apparently, our method is more efficient and appropriate for simultaneous optimization of all network parameters.

After our proposed network is trained, we can perform smoke segmentation on test datasets. We deal with synthetic and real datasets in different ways. Synthetic smoke images have binary ground truths that can be directly computed from alpha channels of pure smoke images. For the sake of convenient comparisons, we perform a binary process on the results for synthetic smoke images, since all comparing methods perform binary segmentation. The binary process regards a prediction value greater than 0.5 as smoke with label "1", and a value less than 0.5 as background with label "0". For real smoke images, we directly retain predicted results without any process for visual accessments because there is no ground truth.

### C. Performance comparisons

Since there is no algorithm for smoke segmentation using deep learning, in order to illustrate performance of the proposed method, we compared the experimental results with multiple classical segmentation algorithms based on deep learning, including FCN [14], SegNet [29], Text-Block FCN [31] denoted as TBFCN, and static map detection method [16] denoted as SMD. We used the network code of the authors and trained these comparing methods on the same smoke training data. In comparisons, there are several points that need to be explained. First, we only compared the first path segmentation results of TBFCN because the method was designed for text segmentation and the work of the second path was proposed to locate the text center point, and it is not needed for smoke segmentation. Second, the method in [16] was proposed for video object segmentation and used dynamic information. For fair comparisons, we only used the static segmentation results of this method for comparisons.

**(1) Qualitative comparisons**

Qualitative comparisons of our method with other deep learning methods on the smoke test datasets are shown in Fig. 5. The first column shows test images, the second column shows corresponding ground truths, and other columns show the segmentation results of different methods. Fig. 6 shows experimental results of real smoke images collected from the internet. Except that the real smoke images do not have corresponding ground truths, the other columns in Fig. 6 are the same as those in Fig. 5. As we can see, our method exhibits excellent segmentation performance on both synthetic smoke images and real smoke images. Indeed, our method separates smoke with sharper edges and more accurate locations compared to other methods.

**(2) Quantitative analysis**

We performed quantitative evaluation experiments on the three synthetic smoke test datasets (DS01, DS02 and DS03).

We calculated two widely used measures: mean Intersection over Union (mIoU) and Mean Square Error (Mse). The larger the value of mIoU is, the better the segmentation is, while the smaller the value of Mse is, the better the result is.

The value of mIoU can well reflect the accuracy of segmentation results. The mean of Intersection over Union (mIoU) is defined as

$$\text{mIoU} = \frac{1}{n} \sum_{i=1}^{n} \frac{PR_i \bigcap GT_i}{PR_i \bigcup GT_i} \qquad (6)$$

where $PR_i$ is the predicted segmentation result of the $i$th image, and $GT_i$ is corresponding ground truth, and $n$ is the number of images in a dataset.

$\text{Mse}_i$ is defined as the average per-pixel square difference between the prediction result $PR$ and its ground truth $GT$ for the $i$th test image:

$$\text{Mse}_i = \frac{\sum_{k=1}^{h_i \times w_i} (PR(x_k) - GT(x_k))^2}{h_i \times w_i} \qquad (7)$$

where $h_i$ and $w_i$ are the height and width of the $i$th test image, and $x_k$ is the coordinates of the $k$th pixel in the $i$th test image. In our experiments, we compute the average Mse on a test dataset for performance evaluation:

$$\text{mMse} = \frac{1}{n} \sum_{i=1}^{n} \text{Mse}_i \qquad (8)$$

The quantitative results on the three test datasets are given in Table 4 to Table 6. As we can see, our method significantly outperforms other methods on all synthetic datasets. Our method achieves the highest mIoU among all the comparing methods, indicating that our prediction segmentation is the closest to its ground truth. At the same time, our method achieves the lowest mMse among them.

Table 4.  Comparisons on DS01

| Methods | mIoU(%) | mMse |
|---|---|---|
| FCN-8s[14] | 64.03 | 0.3221 |
| SegNet[29] | 56.94 | 0.3976 |
| Static Map Detection[16] | 62.88 | 0.3209 |
| Text-Block FCN [31] | 66.67 | 0.3021 |
| Our method | **71.04** | **0.2745** |

Table 5.  Comparisons on DS02

| | mIoU(%) | mMse |
|---|---|---|
| FCN-8s[14] | 63.28 | 0.3353 |
| SegNet[29] | 56.77 | 0.4173 |
| Static Map Detection[16] | 61.50 | 0.3379 |
| Text-Block FCN[31] | 65.85 | 0.3196 |
| Our method | **70.01** | **0.2894** |

Table 6.  Comparisons on DS03

| Methods | mIoU(%) | mMse |
|---|---|---|
| FCN-8s[14] | 64.38 | 0.3202 |
| SegNet[29] | 57.18 | 0.4047 |
| Static Map Detection[16] | 62.09 | 0.3255 |
| Text-Block FCN[31] | 66.20 | 0.3070 |
| Our method | **69.81** | **0.2861** |

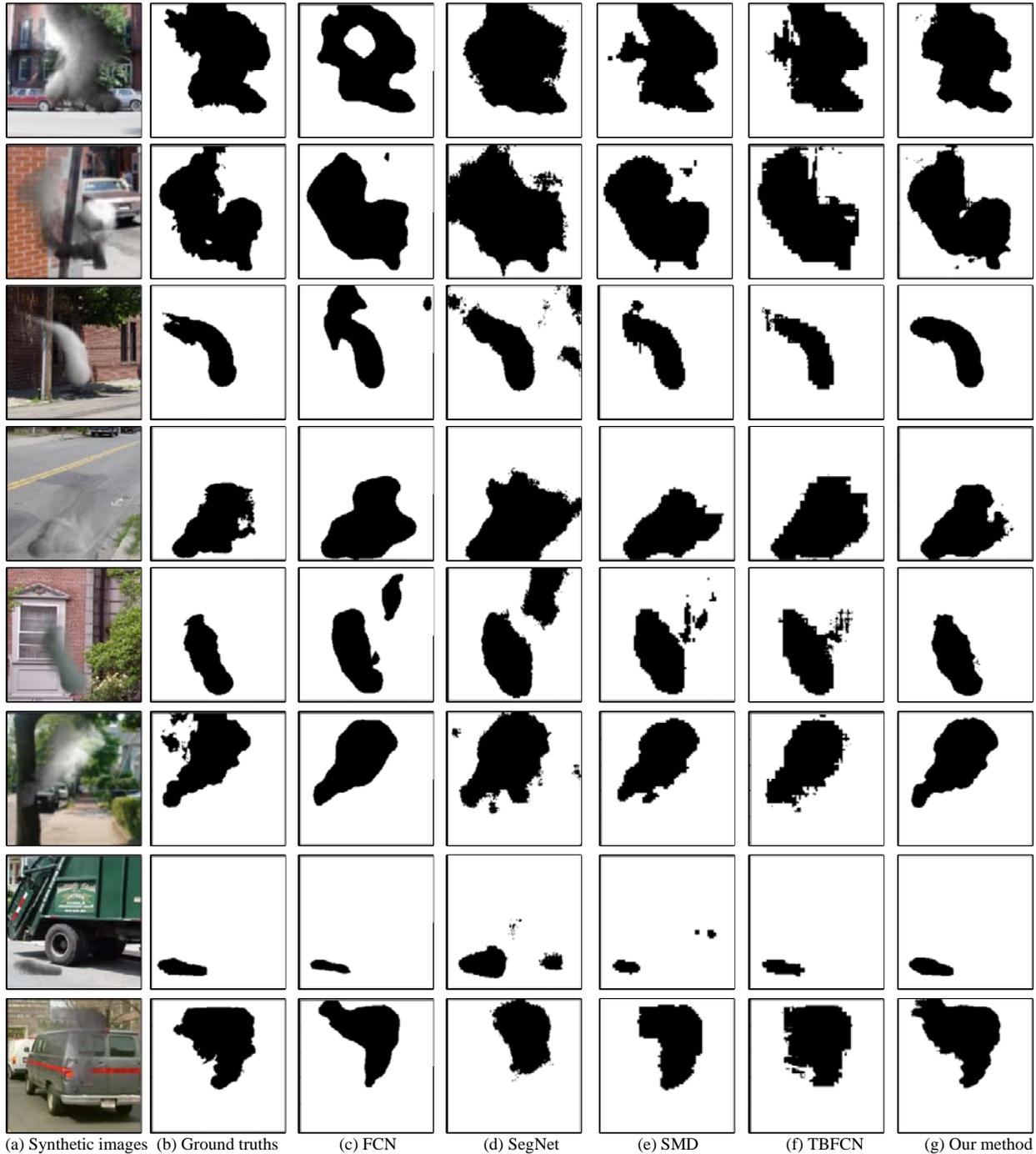

(a) Synthetic images    (b) Ground truths    (c) FCN    (d) SegNet    (e) SMD    (f) TBFCN    (g) Our method

Fig. 5.Segmentation results of synthetic smoke images

Moreover, we have found that the performance of SegNet is the lowest among them. We think that there are two main reasons. The first reason is that SegNet does not adopt the same skip layers as other methods. The second one is that SegNet does not use the pre-trained weights of VGG16 to initialize the network. Network initialization using pre-trained parameters on large image datasets is very favorable in the case of relatively few training samples. Therefore, we can draw some conclusions that skip architectures and pre-trained weights can greatly improve the prediction performance of the proposed model.

*D. Ablation analysis*

In order to analyze the importance and necessity of each part of the proposed network, we conducted a series of ablation experiments. We compared the proposed network with three variants of the proposed network by removing the skip structures of path 2 (-Rs), the entire network of path 2 (-R), and the entire network of path 2 and the skip structures of path 1 (-R-Cs). Ablation experiments will show that skip structures and the second shallow encoder-decoder network play a very important role in smoke segmentation.

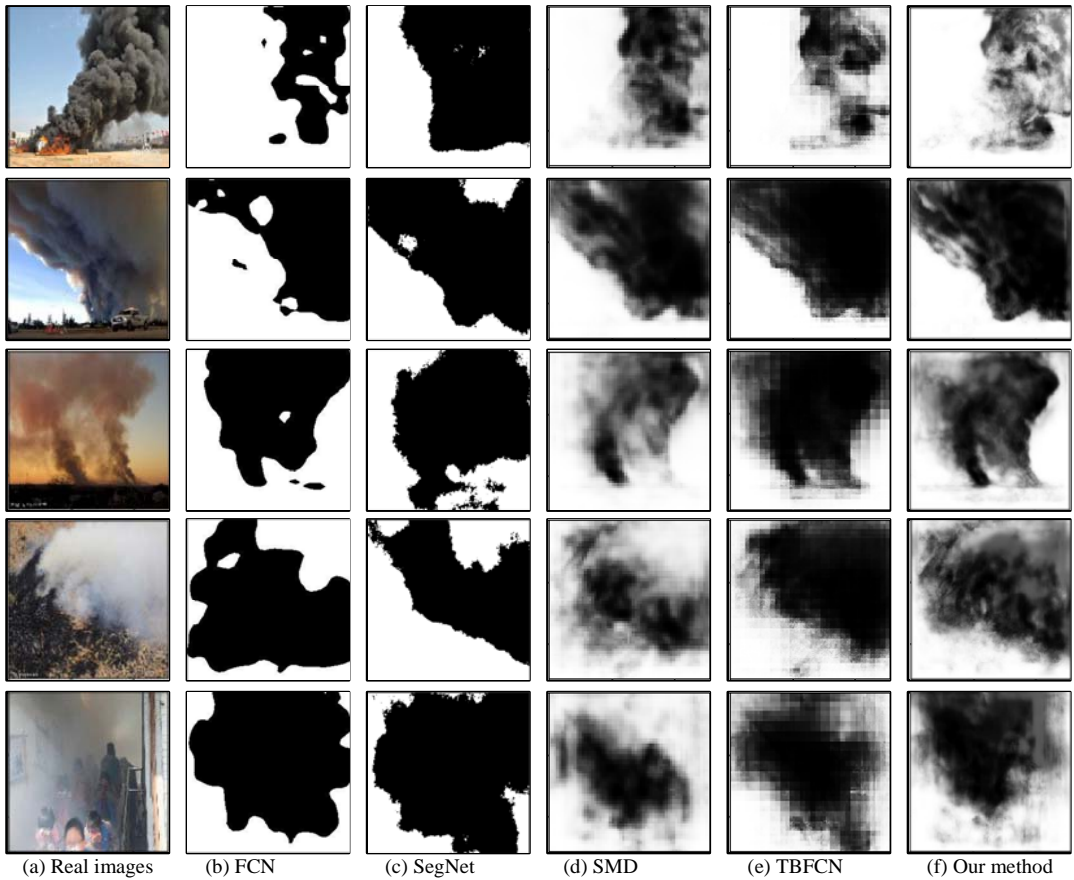

| (a) Real images | (b) FCN | (c) SegNet | (d) SMD | (e) TBFCN | (f) Our method |

Fig. 6.    Segmentation results of real smoke images

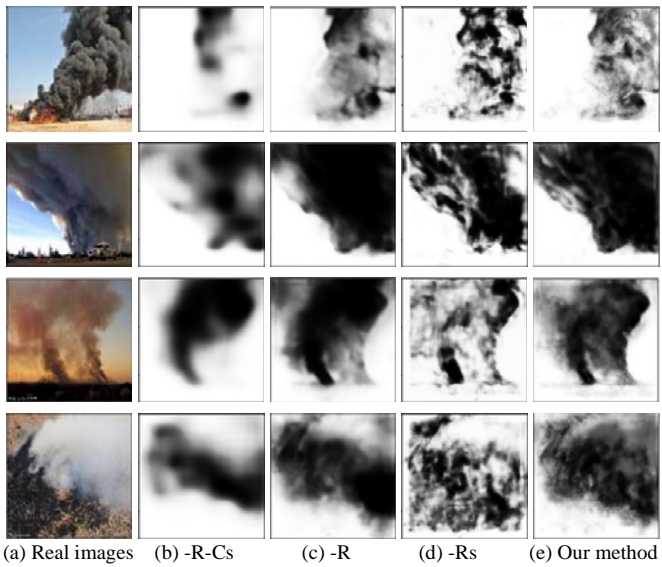

| (a) Real images | (b) -R-Cs | (c) -R | (d) -Rs | (e) Our method |

Fig. 7.    Qualitative comparisons of ablation experiments on real images

We visually evaluated the results of the three variants and the proposed complete method on real images. Qualitative experimental results are shown in Fig. 7. From Fig. 7, we can clearly observe that the final results of the proposed method are significantly better than the coarse results of Variant 2 (-R).

Meanwhile, the networks with skip layers in the two paths, i.e. Variant 2 (-R) and our complete method, are better than the methods without skip layers that are Variant 1 (-R-Cs) and Variant 3 (-Rs).

Table 7.  Ablation results on DS01

|                   | mIoU  | mMse   |
| ----------------- | ----- | ------ |
| Variant 1 (-R-Cs) | 61.71 | 0.3387 |
| Variant 2 (-R)    | 64.64 | 0.3033 |
| Variant 3 (-Rs)   | 66.03 | 0.2985 |
| Our method        | **71.04** | **0.2745** |

Table 8.  Ablation results on DS02

|                   | mIoU  | mMse   |
| ----------------- | ----- | ------ |
| Variant 1 (-R-Cs) | 56.87 | 0.3525 |
| Variant 2 (-R)    | 63.97 | 0.3162 |
| Variant 3 (-Rs)   | 64.54 | 0.3092 |
| Our method        | **70.01** | **0.2894** |

Table 9.  Ablation results on DS03

|                   | mIoU  | mMse   |
| ----------------- | ----- | ------ |
| Variant 1 (-R-Cs) | 57.99 | 0.3445 |
| Variant 2 (-R)    | 64.51 | 0.3065 |
| Variant 3 (-Rs)   | 65.57 | 0.3043 |
| Our method        | **69.81** | **0.2861** |

Table 10. Comparison results between two fusion methods on the three test smoke datasets

| | DS01 | | DS02 | | DS03 | |
|---|---|---|---|---|---|---|
| | mIoU | mMse | mIoU | mMse | mIoU | mMse |
| Variant 4 (deconvolution+add) [31] | 67.42 | 0.3032 | 67.28 | 0.3126 | 66.94 | 0.3094 |
| Our method (upsampling+concatenation) | **71.04** | **0.2745** | **70.01** | **0.2894** | **69.81** | **0.2861** |

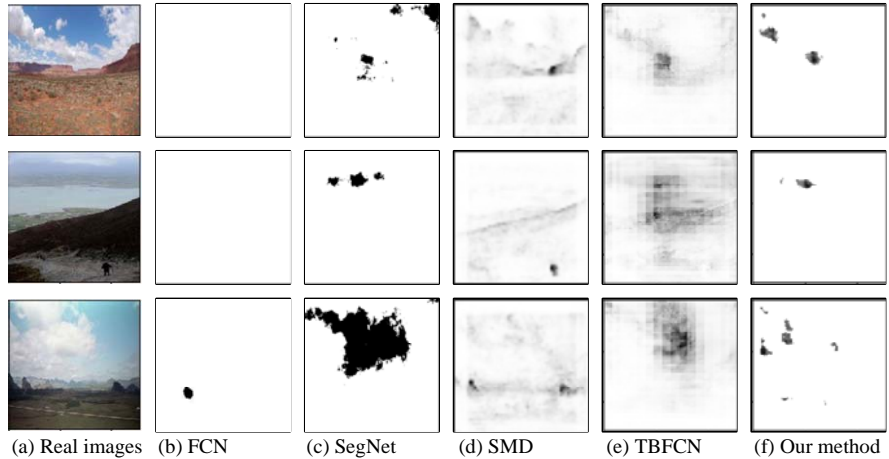

(a) Real images  (b) FCN  (c) SegNet  (d) SMD  (e) TBFCN  (f) Our method

Fig. 8.    Some examples of false positives

We also performed ablation experiments on the three synthetic datasets. Quantitative results are listed in Table 6 to Table 8. According to the quantitative results on DS01, we can find that mIoU increases from 64.64% to 71.04% by adding the network of path 2, so it shows that the second network is very important for improvement of accuracy. In addition, the mIoU of the proposed method with skip architectures in both paths is about 4 to 5 percentage higher than those without skip layers. That proves the effectiveness of skip layers. This phenomenon also happens on DS02 and DS03. Therefore, we can see that skip structures and the second refinement network play a very important role in our network.

Since the method of [31] used deconvolution and add operations to fuse features from different layers, we want to verify that upsampling and concatenation operations used in our proposed method is more appropriate and effective than deconvolution and add operations used in [31]. Therefore, we replace unsampling and concatenation operations by deconvolution and add layers to produce a new variant of the proposed method, denoted as Variant 4 (deconvolution+add). We compared the proposed method with Variant 4 on the three synthetic smoke datasets under the same condition. The experimental results are shown in Table 10. We can find that unsampling and concatenation operations unanimously achieve better performance than deconvolution and add ones on DS01, DS02, and DS03.

*E.  Test on videos*

We tested our method on the same four videos as [46]. The four videos are two smoke videos and two non-smoke videos, respectively. The first video is a black smoke video produced by burning diesel oil. Our method achieved accurate segmentation for each frame, as shown in Fig. 9. The second video is a white smoke video by cotton ropes, as shown in Fig.

10. The white smoke video has poor image quality, leading to inaccurate segmentation results. The third is a non-smoke video containing waving leaves, and the fourth is a basketball court video with some students playing basketball. Fig. 11 shows two frames from the two non-smoke videos. We are pleased that our method did not misclassify any pixel as smoke on the two non-smoke videos, so we do not illustrate segmentation masks.

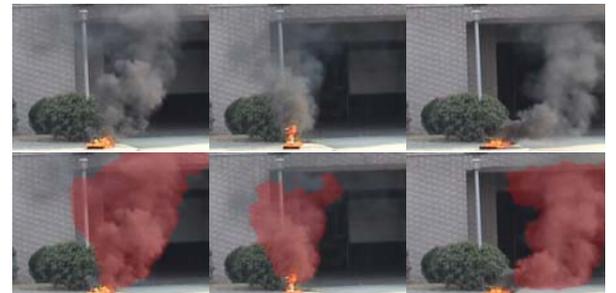

Fig. 9.    Frames and corresponding segmentation masks on black smoke

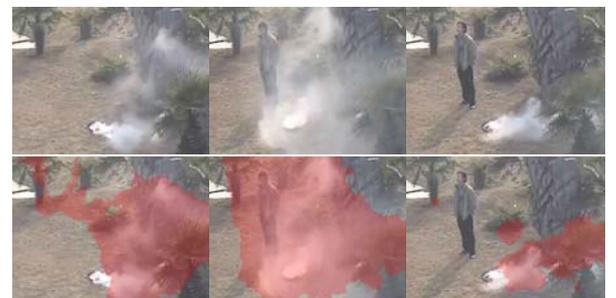

Fig. 10.   Frames and corresponding segmentation masks on white smoke

Based on the segmentation results of our method, we can easily make the whole image smoke recognition. For the sake of simplicity, we simply count the number of pixels classified as smoke in an image, and classify the image as a smoke one if the number of pixels classified as moke is greater than a threshold.

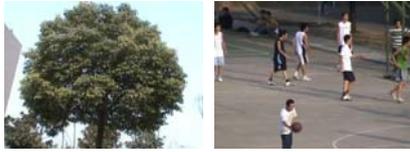

Fig. 11.　Two frames from a waving leaf video (left) and a basketball court video (right), respectively

We used the above segmentation based smoke recognition method to detect smoke on the four videos, and compared our method with LBP_LBPV [46] and Toreyin's method [47]. Table 11 lists the results of the three methods. Our method detected smoke at the first frame of the two smoke videos. Apparently, our method can detect smoke much earlier than other two methods. In addition, our method did not raise any false alarms on the two non-smoke videos, and we do not use any post-processing techniques to further reduce false alarms.

Table 11.Comparison results of smoke detection on videos

| Videos | Duration | LBP_LBPV | Toreyin's method | Our method |
|---|---|---|---|---|
| | | Smoke detected at frame # | | |
| (a) black smoke | 517 | 89 | 164 | 1 |
| (b) white smoke | 2886 | 94 | 216 | 1 |
| | | Number of false alarms | | |
| (c) waving leaves | 895 | 0 | 0 | 0 |
| (d) basketball court | 4536 | 1 | 4 | 0 |

### F. Limitations of our method

Although our method has achieved good results, there is a long way to reach perfect effects. The main reasons are as follows. First, smoke has very large variations of features: texture, shape, color, etc., and these features appear in many forms and are also varying even if they are produced from the same fire source. Second, the edge of smoke is very blurry compared to other objects, and smoke precise edges are hard to get. Third, many objects, such as fog and clouds, share the same visual appearance as smoke. It is very difficult to discriminate them. As shown in Fig. 8, there are several examples of false positives produced by the comparing methods and the proposed method. Except for FCN, our method only misclassified a small number of fog and cloud pixels as smoke compared to other methods. This also proves that our proposed method not only has certain advantages in accuracy, but also has a strong competitive effect on false positives. Moreover, real-time performance is a very important requirement for application of smoke segmentation. Therefore, another limitation of our method is that the test speed needs to be further improved.

## VI. Conclusions

In recent years, fully convolutional networks (FCN) have exhibited outstanding performance in semantic segmentation. Inspired by the success of FCNs, we propose a two-path encoder-decoder FCN to solve the difficulty of visual smoke segmentation. Our proposed network consists of coarse and fine prediction paths. The network of the first path is an encoder-decoder FCN architecture with skip structures for extraction of global information, while the second path is an encoder-decoder FCN framework with skip structures, but it is significantly shallower, smaller and simpler than the first path network. The two networks work together to solve extraction of smoke context information and localization features. The fuzzy attribution of smoke directly leads to the difficulty in manually annotating smoke images, so we use an image composition method to generate synthetic smoke images. We performed experiments on three synthetic smoke datasets and a realistic smoke dataset collected from the internet. Comparison experiments show that our method truly achieves much better performance than the state-of-the-art segmentation algorithms based on FCNs. Ablation experiments also verify that the second refinement network and skip structures are very important for improvement of smoke segmentation accuracy. Experiments on videos show that our method also outperforms existing methods.

### Acknowledgments

This work was partially supported by the National Natural Science Foundation of China (61862029), and the Science Technology Application Projects of Jiangxi Province (KJLD12066, GJJ170317).